# Amplifying Sine Unit: An Oscillatory Activation Function for Deep Neural Networks to Recover Nonlinear Oscillations Efficiently

**Jamshaid Ul Rahman**[1,2,*]
[1]*School of Mathematical Sciences, Jiangsu University 301 Xuefu road, Zhenjiang 212013, China.*
jamshaidrahman@gmail.com

**Faiza Makhdoom**[2]
[2]*Abdus Salam School of Mathematical Sciences, GC University, Lahore 54600, Pakistan*
makhdumfaiza@gmail.com

**Dianchen Lu**[1]
[1]*School of Mathematical Sciences, Jiangsu University 301 Xuefu road, Zhenjiang 212013, China.*
dclu@ujs.edu.cn

**Abstract**

Many industrial and real life problems exhibit highly nonlinear periodic behaviors and the conventional methods may fall short of finding their analytical or closed form solutions. Such problems demand some cutting edge computational tools with increased functionality and reduced cost. Recently, deep neural networks have gained massive research interest due to their ability to handle large data and universality to learn complex functions. In this work, we put forward a methodology based on deep neural networks with responsive layers structure to deal nonlinear oscillations in microelectromechanical systems. We incorporated some oscillatory and non-oscillatory activation functions such as growing cosine unit known as GCU, Sine, Mish and Tanh in our designed network to have a comprehensive analysis on their performance for highly nonlinear and vibrational problems. Integrating oscillatory activation functions with deep neural networks definitely outperform in predicting the periodic patterns of underlying systems. To support oscillatory actuation for nonlinear systems, we have proposed a novel oscillatory activation function called Amplifying Sine Unit denoted as ASU which is more efficient than GCU for complex vibratory systems such as microelectromechanical systems. Experimental results show that the designed network with our proposed activation function ASU is more reliable and robust to handle the challenges posed by nonlinearity and oscillations. To validate the proposed methodology, outputs of our networks are being compared with the results from Livermore solver for ordinary differential equation called LSODA. Further, graphical illustrations of incurred errors are also being presented in the work.

**Keywords:** Amplifying Sine Unit, Neural Networks, Microelectromechanical Systems, Nonlinear Dynamics, Non-Monotonic, Oscillatory Activation Function,

## 1 Introduction

Nonlinear problems are pervasive in various fields of science and technology. Many scientific and industrial systems involve components which have complex nonlinear and periodic interactions among them. Such systems can be modeled as differential equations (DEs) carrying the information about their periodicity and nonlinearity. One of the major challenges is to provide accurate analytical solutions of these strongly nonlinear and periodic DEs with computational efficiency. Therefore, modeling and simulation of dynamical systems [1, 2, 3] require critical mathematical thinking and sophisticated computational tools to simulate their solutions. Recently, many semi-analytic iterative methods [4, 5, 6] have been put forward by the researchers to address the nonlinear oscillatory behaviors of underlying systems. The deep neural networks (DNNs) have grown as the most effective architectures [7] to deal with complex patterns and relationships present in data making them well-suited for systems with various interacting components. The hierarchical structure [8] help DNNs to learn increasingly complex patterns or interactions present in data. Depending upon the nature of problem being handled by DNNs, the choice of appropriate activation function is crucial [9]. Linear behaviors might easily be predicted by stacking up some layers of network which are just collecting the linear combinations from previous layers as (1.1)

$$a_i = \sum_i w_i x_i + b_i \qquad (1.1)$$

Therefore, using linear activation functions for a network is not worthy enough as their role can easily be replaced by a single linear layer network shown in Fig. 1.

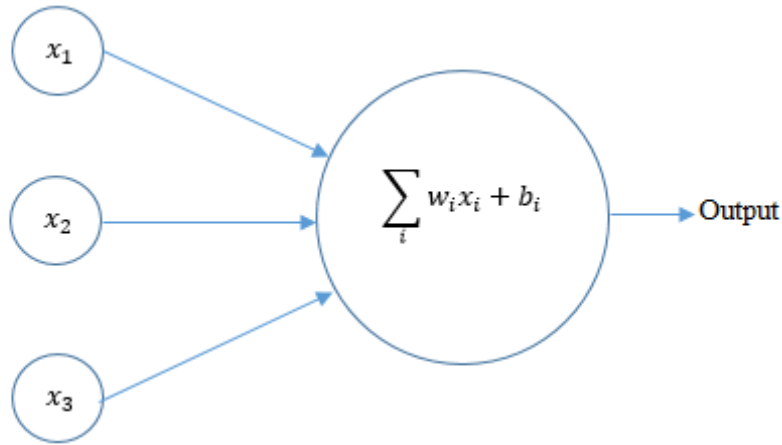

*Fig. 1   A single layer neural network which can only perform linear operation. Such network can only handle linear problems or alternatively can help drawing linear decision boundaries*

Nonlinear activation functions play a significant role to catch the nonlinearity present in data being assigned to DNNs for training. Monotonic nonlinear activation functions [10] such as Sigmoid [11], Tanh [12] and especially ReLU [13] are being abundantly used by the researchers and industry. Despite their significant contribution towards many scientific applications, these activation functions may appear problematic during backpropagation [14] when the parameters are being updated using negative gradients. For instance, sigmoidal activations undergo vanishing gradient problems [15] and ReLU based networks may suffer dying ReLU problem [16] which may result in decreased performance and slower convergence. These gradient based issues during backpropagation can be avoided by the utilization of SELU [17] and ELU [18] which are the variants of ReLU. Till recent past, all the research was based on the development of non-oscillatory and monotonic activation functions. Swish [19] and Mish [20] introduced the effective and powerful use of non-monotonic functions as activations of DNNs for various scientific and industrial problems [21, 22]. GCU [23], an oscillatory non-monotonic activation function revolutionized the domain of activation functions by breaking the custom of utilizing only non-oscillatory activations. GCU provided the single neuron solution to XOR problem as a neuron with oscillatory activation can individually perform nonlinear decisions. Thus, oscillatory activation functions [24] are advantageous to perform the complex assigned tasks with lesser number of neurons and are computationally cheaper with enhanced performance.

Other than activation functions, the training and efficiency of DNNs can also be influenced by the right choice of loss functions and optimization algorithms [25, 26] depending upon the demand of problem. For instance, recognition and classification tasks in the field of computer vision [27] may work well with the modifications of softmax loss [28, 29] including Sphereface and additive parameter approaches [30, 31, 32]. On the other hand, the task of approximating the solutions to DEs using neural networks can be modeled well by applying simple squared loss functions [33, 34]. We are limiting our discussion to just elaborate the role of different oscillatory and non-oscillatory activation functions to deal highly nonlinear and periodic systems using DNNs.

A new oscillatory activation function ASU given by (1.2) has also being proposed and its behavior is graphically presented by Fig. 2.

$$ASU(a) = a.\sin(a) \qquad (1.2)$$

The ASU is non-monotonic and as input gets larger, its oscillations tend to amplify.

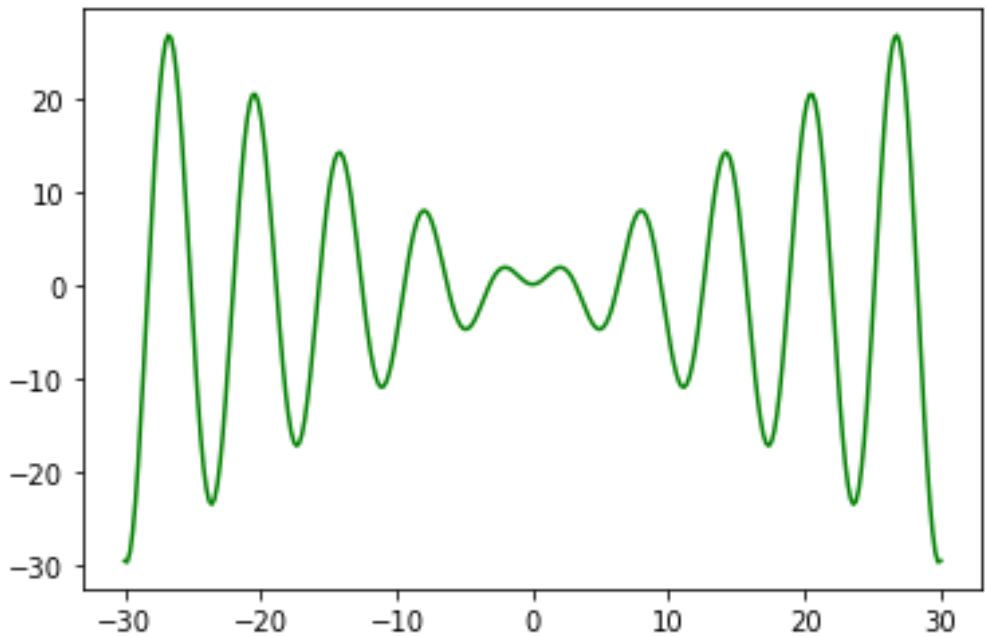

Fig. 2:   *Graphical illustration of proposed activation function ASU that efficiently deal with nonlinear periodic problems*

We have designed DNNs with modification in [35] using different activation functions for highly nonlinear oscillatory systems [36, 37]. Outputs of DNNs are being compared with numerical solutions of LSODA routines [38] offered by a FORTRAN77 library [39] which include Adams-Bashforth [40] and BDF [41] methods. The implementations are being executed by PyTorch [42, 43] and numerical simulations performed by SciPy [44] Odeint library [45].

## 2 Methodology

This section covers the detailed methodology adopted to capture complex nonlinear oscillations. Modifications in DNNs on the basis of oscillatory and non-oscillatory activation functions are summarized in the following sub sections.

## 2.1 Formulation of DNN

The strategy is to first mathematically model the problem as a differential equation i.e.

$$Lf - p = 0 \qquad (2.1)$$

where $L$ is the differential operator, $f$ is the dependent variable wished to be explored and $p$ represents a general expression for the remaining terms involved in the mathematical model. If $f_N$ is the solution by DNN then it should have to satisfy (2.1) with the imposed initial or boundary conditions. To fulfill the criteria of satisfying the subjected conditions, $f_N$ can set to undergo the following transformation

$$\tilde{f} = f_0 + (1 - e^{-(t-t_0)})f_N \qquad (2.2)$$

where $f_0$ is the imposed initial condition and (2.2) is one of the choices made to ensure the satisfaction of subjected conditions. Next step for $\tilde{f}$ to be a valid solution is that the residual $L\tilde{f} - p$ should identically be zero. The loss function given by (2.3) is formulated to train the network so that after a thorough optimization process the residuals are as minimized as possible.

$$Loss = (L\tilde{f} - p)^2 \qquad (2.3)$$

The detailed design and implementation of our DNN models are discussed in section 4.

## 2.2 Activation functions taking part in DNN based approximations

In order to recover the dynamics of highly nonlinear systems which undergo periodic motions the choice of oscillatory activation functions is more robust, feasible, time and cost saving. It is still possible to use non-oscillatory activation functions for such problems but it would be more prone to high computational cost, training loss and time consumption. To validate our stance, we analyzed the behaviors of five different activation functions for a highly nonlinear oscillatory system.

**Table 1** A mathematical representation and nature of different activation functions

| Name | Function | Nature |
|---|---|---|
| Tanh | $f(z) = \dfrac{e^z - e^{-z}}{e^z + e^{-z}}$ | Non-oscillatory and monotonic |
| Mish | $f(z) = z.\tanh(\log(1 - e^z))$ | Non-oscillatory and non-monotonic |
| Sine | $f(z) = \sin(z)$ | Oscillatory and non-monotonic |
| GCU | $f(z) = z.\cos(z)$ | Oscillatory and non-monotonic |
| ASU | $f(z) = z.\sin(z)$ | Oscillatory and non-monotonic |

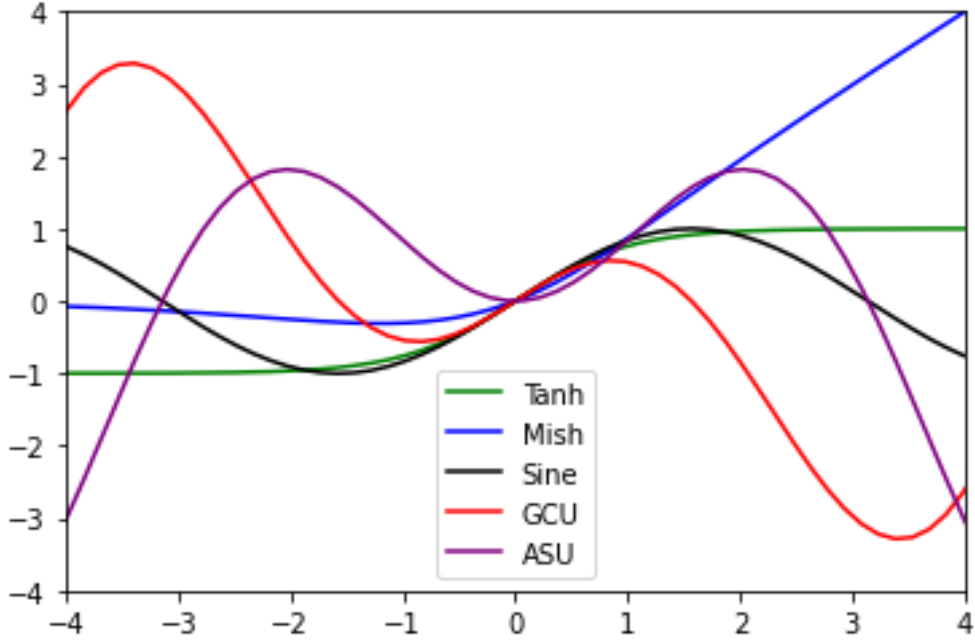

**Fig. 3** *Plots of different oscillatory and non-oscillatory activation functions for nonlinear periodic systems*

Table 1 provides the detail of activation functions and their nature which have a significant effect on their performance and Fig. 3 provides the graphical illustrations of proposed ASU and other activation functions. It can be observed that Tanh is monotonic, where as ASU, GCU and Sine are oscillatory and non-monotonic. On the other hand, Mish is non-oscillatory and non-monotonic.

## 3 Application in MEMS

Microelectromechanical systems (MEMS) can both be referred to a technology as well as the miniature, complex, sophisticated and reliable devices manufactured on small silicon chips using microfabrication techniques. In recent years, MEMS have gained a massive research interest as they have become an inevitable part of every modern industry. Inherently, MEMS show highly nonlinear periodic behaviors and the classical approaches may fall short to deal these sensitive microstructures. Therefore, MEMS need some state-of-the-art, carefully devised and efficient techniques to be tackled by. This paper discusses the utilization of DNNs with different activations to deal highly nonlinear and periodic behaviors in MEMS.

### 3.1 Electrically actuated clamped-clamped MEMS

Suppose a clamped-clamped MEMS with a microbeam of length $l$, width $b$, thickness $h$ and density $\rho$ as shown in Fig. 4. The system is actuated through an electrostatic force given by (3.1)

$$F_e = \frac{bV^2 \varepsilon_V}{2} \left[ \frac{1}{(d-W)^2} - \frac{1}{(d+W)^2} \right] \quad (3.1)$$

where $V$ denotes the applied voltage, $\varepsilon_V$ is dielectric constant and $d$ is the initial gap between substrate and beam. The deflection of beam in electrically actuated MEMS [46] can be modeled as

$$(A_0 + A_1 u^2 + A_2 u^4)\frac{d^2 u}{dt^2} + A_3 u + A_4 u^3 + A_5 u^5 + A_6 u^7 = 0 \tag{3.2}$$

subjected to initial conditions

$$u(0) = \frac{\pi}{3}, \qquad u'(0) = 0 \tag{3.3}$$

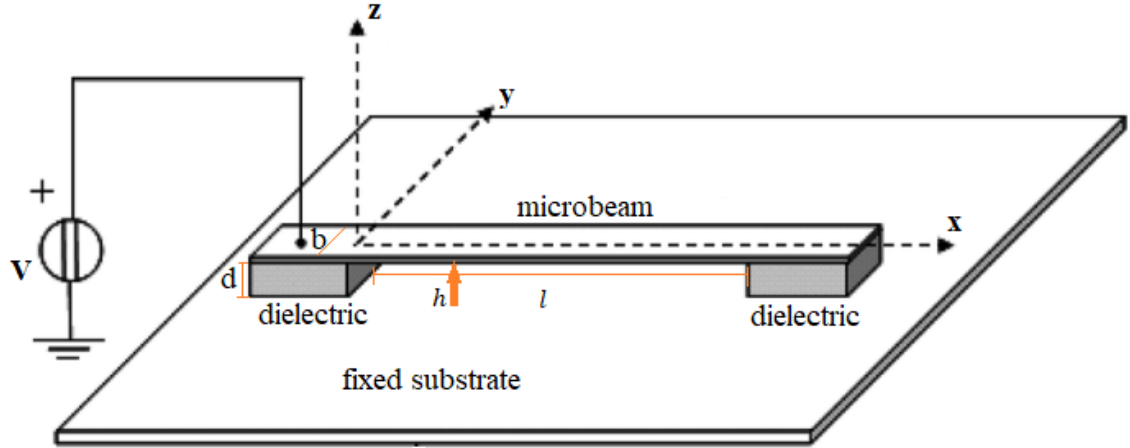

**Fig. 4** *Clamped-clamped microbeam MEMS undergoing electrostatic force based actuation where d is the initial gap between the beam and fixed substrate*

## 4 Implementation

The designed DNNs (utilizing activation functions discussed in Table 1) consist of three hidden layers with 128 neurons each and one output unit which spits out $\tilde{f}$. In order to train the network for our modeled MEMS, the squared loss is being calculated using (2.3) and networks' weights are adjusted using Adam as an optimizer. Implementations are being executed by PyTorch and Fig. 5 hierarchically defines the steps involved in DNN-based approximations.

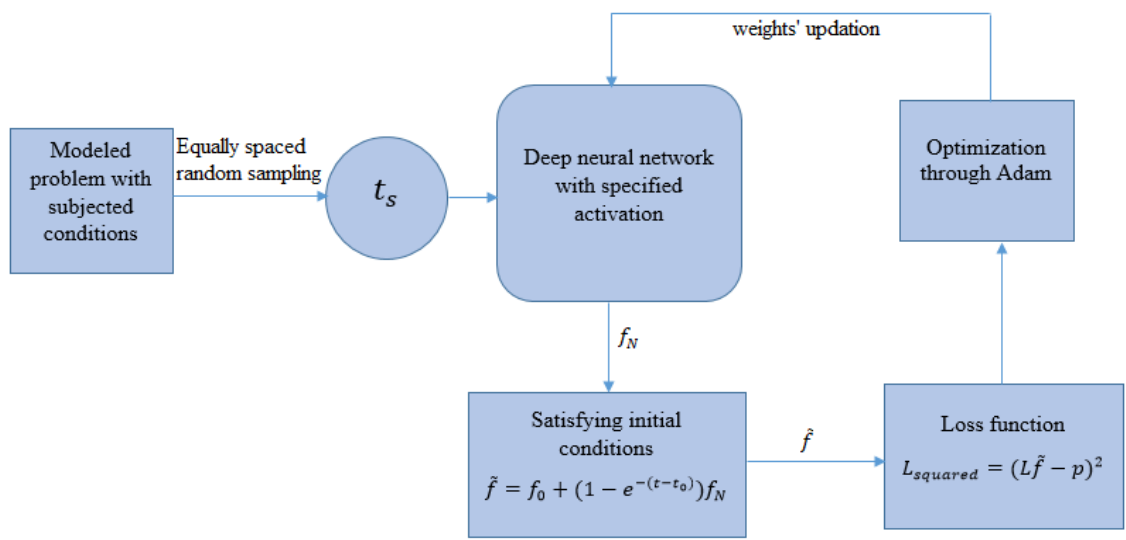

**Fig. 5** *Flowchart representation of the methodology followed to simulate the dynamics of MEMS using DNNs*

Outputs of DNNs are then being compared with the numerical simulations performed in SciPy. In order to have a comprehensive view of overall performance, an error analysis for each simulation is also being performed. For a given time domain, the error values simply represent the difference between the solution values approximated by DNNs and other numerical techniques.

## 5 Results and discussion

This section provides a detailed discussion on the outcomes of DNNs for electrically actuated MEMS. The performance of networks for each activation function has been discussed case by case.

### 5.1 DNN-based Simulations with non-oscillatory activation functions

We performed two experiments to recover the dynamics of MEMS modeled mathematically by (3.2) and (3.3). First experiment utilizes Tanh activations throughout the hidden layers and the other one is with Mish as an activation function. The training of DNNs with non-oscillatory activation functions needed large number of epochs, thus was slow and time consuming.

#### 5.1.1 Utilization of Tanh

The network with Tanh activations was able to approximate the solution to the underlying MEMS after a training process of 35,000 epochs. Fig. 6(a) provides the solution curve approximated by DNN with Tanh activations.

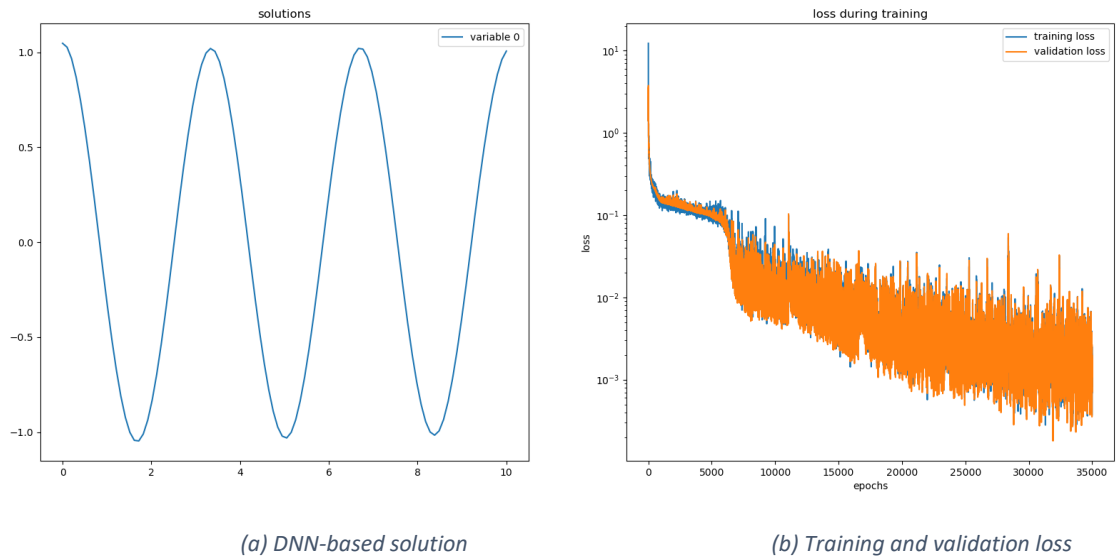

(a) DNN-based solution          (b) Training and validation loss

*Fig. 6*     *The recovered solution and loss history in case of Tanh as an activation function for the network*

Fig. 6(b) provides the history of both the training and validation loss. The lowest calculated loss is 0.00017960761963920115 and it took 18 minutes and 33 seconds to accomplish the training and get the solution. Fig. 7(a) illustrates the comparison of DNN-based and numerical solutions and Fig. 7(b) demonstrates the incurred error in the specified time span.

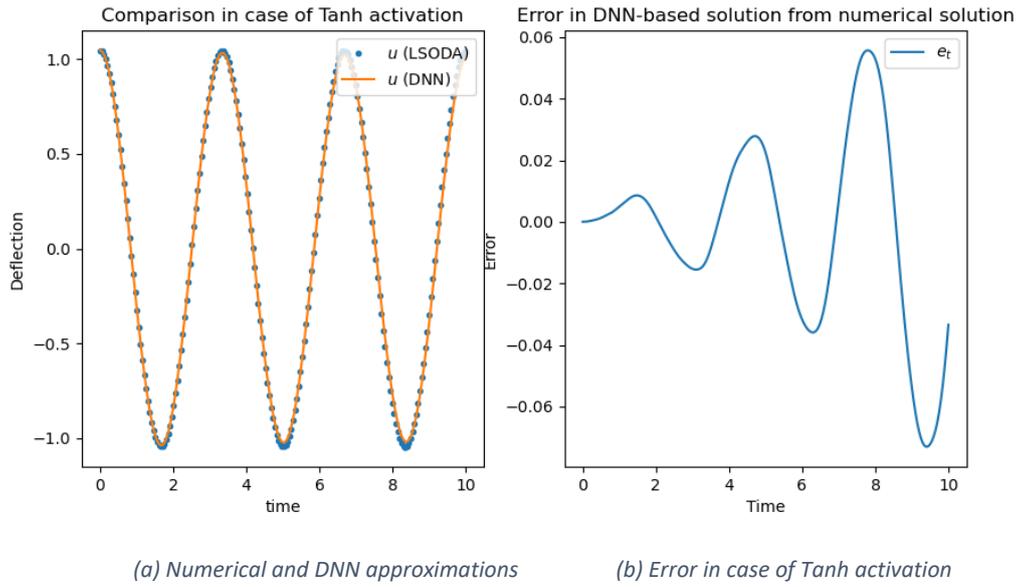

(a) Numerical and DNN approximations    (b) Error in case of Tanh activation

**Fig. 7**    Comparison and error illustrations of solutions approximated by both the DNNs and numerical solvers

The error simply represents difference between the solutions approximated by our designed networks and numerical simulations.

### 5.1.2 Utilization of Mish

The network using Mish as an activation function recovered the solution after 20,000 training epochs and this process took 17 minutes and 50 seconds. Fig. 8(a) presents the resultant solution of MEMS and Fig. 8(b) provides the loss history. The lowest calculated loss in this case is 0.00015319963267778283.

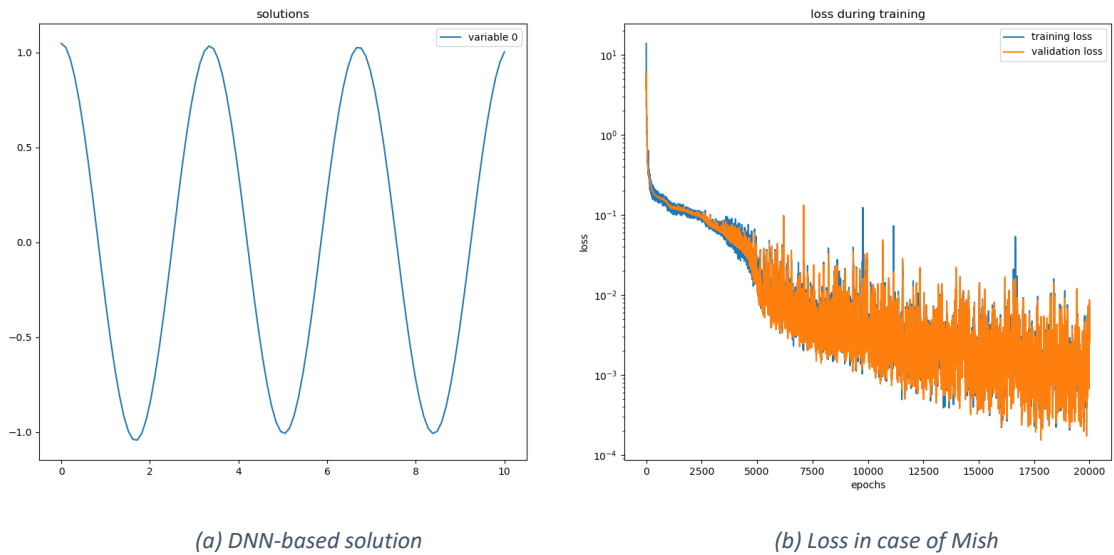

(a) DNN-based solution    (b) Loss in case of Mish

**Fig. 8**    DNN outcomes for the approximation of deflection of electrically actuated microbeam in MEMS

Fig. 9(a) depicts the comparison of outputs and Fig. 9(b) graphically presents the error between DNN and LSODA based approximations.

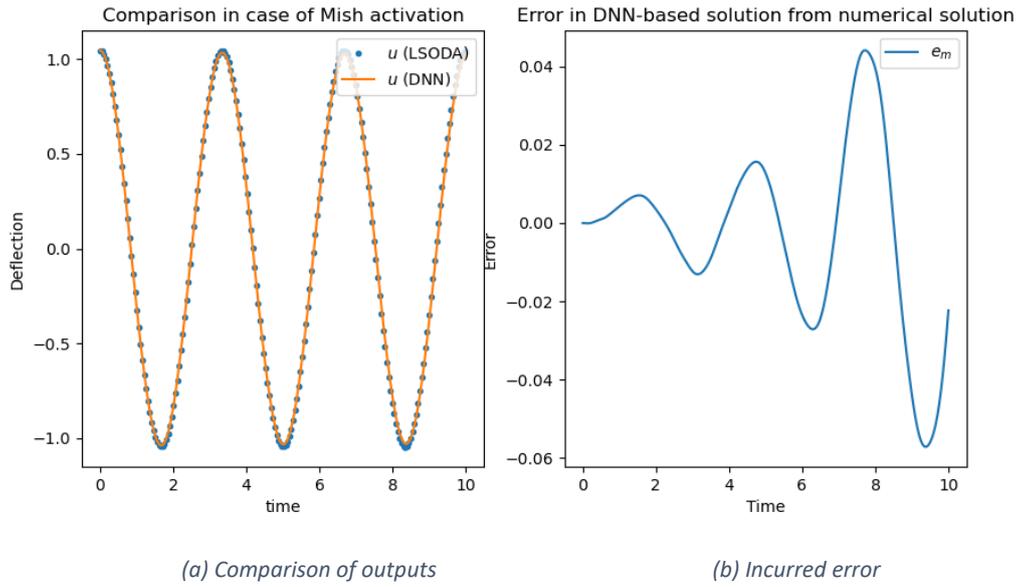

(a) Comparison of outputs  (b) Incurred error

**Fig. 9**  Graphical comparison of solutions from both the techniques and discrepancy between both the approximations

### 5.2 DNN-based Simulations with oscillatory activation functions

This section discusses the utilization of oscillatory activation functions to approximate the solutions of nonlinear periodic dynamical systems. The training of DNNs with oscillatory activation functions was much faster and efficient.

### 5.2.1 Deployment of Sine

In case of Sine activations, DNN approximated the solution after a training of 6,000 epochs. The training was much faster than the case of non-oscillatory activation functions as it completed in 3 minutes and 16 seconds.

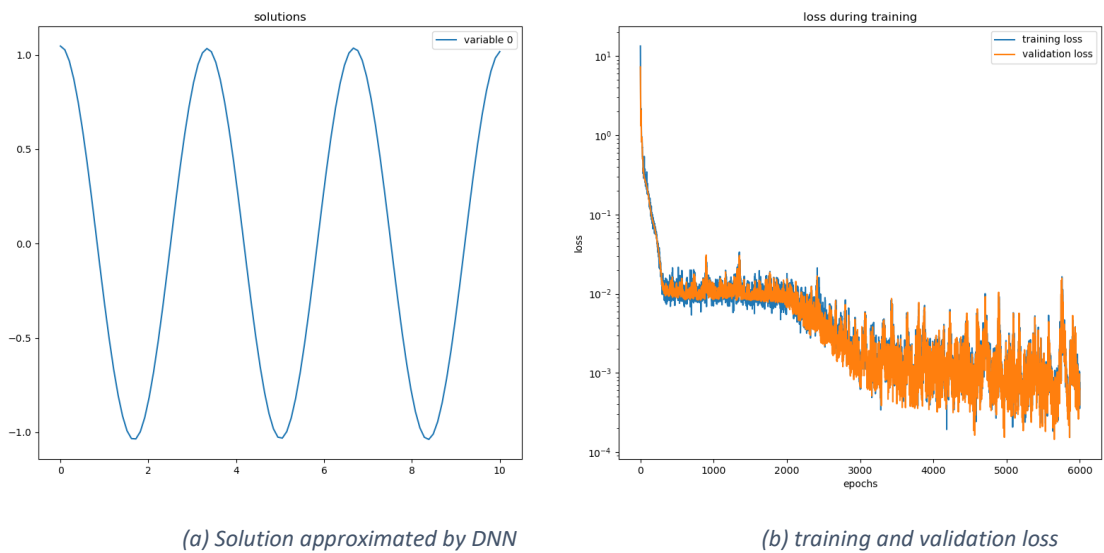

(a) Solution approximated by DNN  (b) training and validation loss

**Fig. 10**  DNN-based simulation results and loss history in case of Sine as an activation function

Fig. 10(a) provide the illustrations of recovered solution after training and Fig. 10(b) details the training and validation loss. The lowest calculated loss is

0.00014502210832914825. Fig. 11(a) presents the comparison of solutions from DNN and LSODA solvers while Fig. 11(b) illustrates the corresponding error between their values.

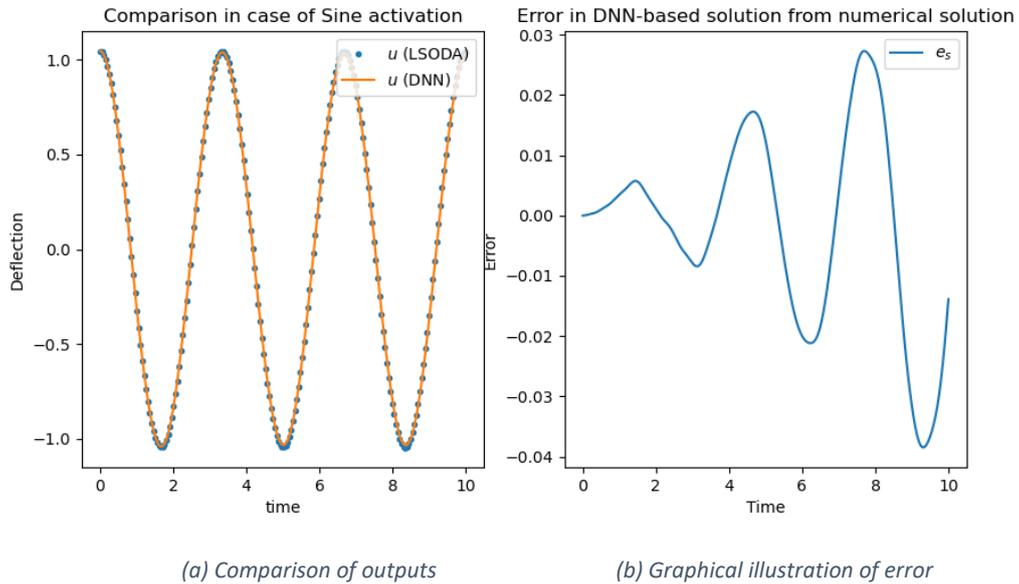

*(a) Comparison of outputs*   *(b) Graphical illustration of error*

**Fig. 11** Comparison of approximations for the deflection of beam and error calculations for the case of Sine based DNN

One can observe that the error fairly small when the network used an oscillatory activation i.e. Sine to approximate the solution of a highly nonlinear periodic problem. In other words it can be said that Sine, unlike non-oscillatory functions, has enhance the ability of DNN to predict the oscillations in MEMS.

### 5.2.2 Deployment of GCU

When the hidden layers of DNN were implemented by GCU as an activation function, the efficiency of network enhanced remarkably. The network was able to predict the solution after a training of just 4,000 epochs and the process took 2 minutes 53 seconds. Fig. 12(a) represents the solution predicted by the network and loss history is mentioned by Fig. 12(b).

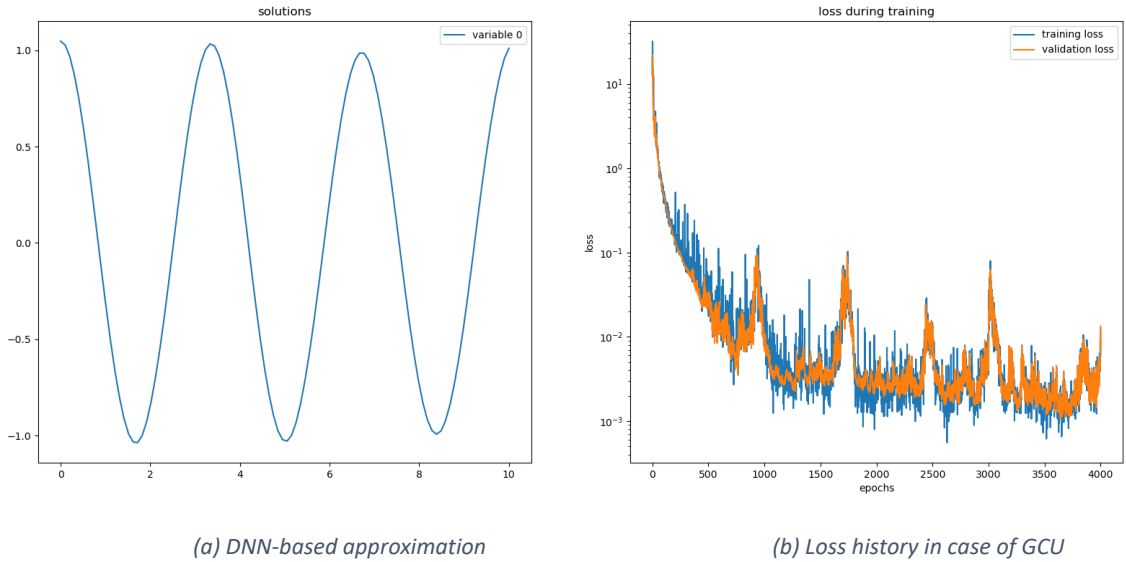

(a) DNN-based approximation

(b) Loss history in case of GCU

**Fig. 12** Output of network and details of training and validation loss in case of GCU activation function

Fig. 13(a) provides the detailed comparison of DNN-based and numerical solutions. The error analysis presented by Fig. 13(b) is also ensuring the capability of GCU to perceive the high oscillatory patterns.

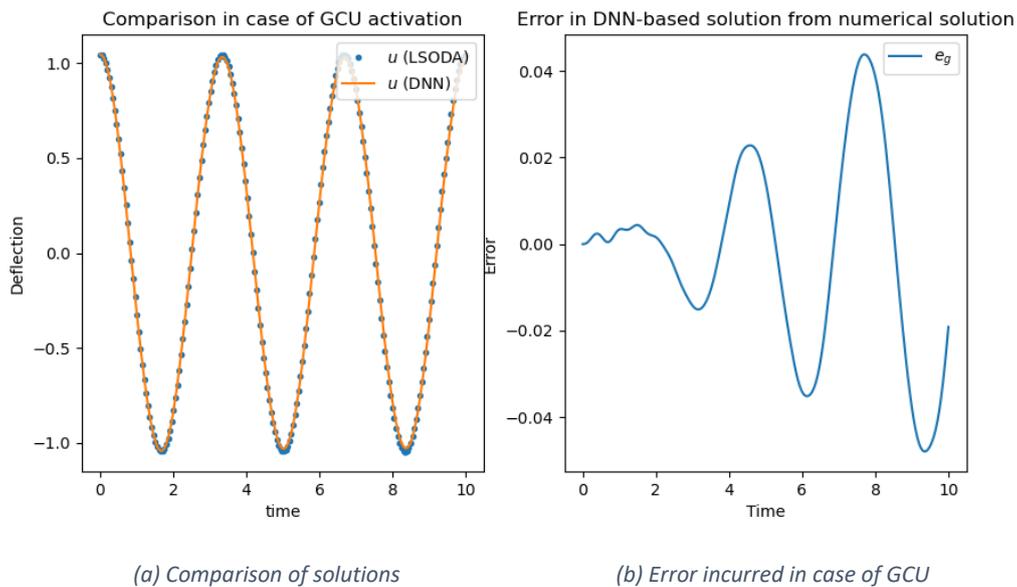

(a) Comparison of solutions

(b) Error incurred in case of GCU

**Fig. 13** Output from DNN being compared by numerical simulations and the error which is representing the difference between the solutions approximated by both the techniques

Now we will discuss the details of last experiment which was performed by utilizing proposed activation function ASU.

### 5.2.3 Deployment of ASU

The DNN architecture with ASU activation function was super-fast and vigilant to discover the nonlinear oscillatory patterns of underlying MEMS. The network approximated the solution efficaciously after a training of just 2,000 epochs and the process completed in a short time of just 1 minute and 14 seconds. Fig. 14(a) provides us with the graphical representation of solution whereas the training and validation loss are presented by Fig. 14(b).

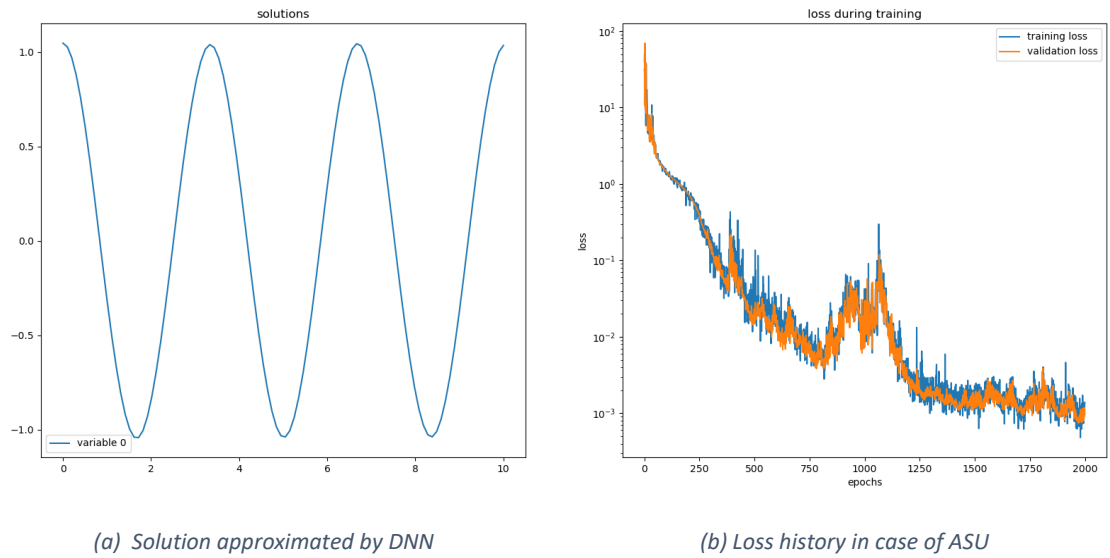

(a) Solution approximated by DNN

(b) Loss history in case of ASU

**Fig. 14** *Illustrations of approximated solution by DNN utilizing ASU as activation along with training and validation loss history*

Fig. 15(a) provides the detailed comparison of DNN-based solution with numerical approximation by LSODA and Fig. 15(b) depicts the error evaluation of the network's output. It can be observed that besides the competence of DNN with ASU to deal nonlinear oscillations rigorously, the error also fairly lie in an acceptable range.

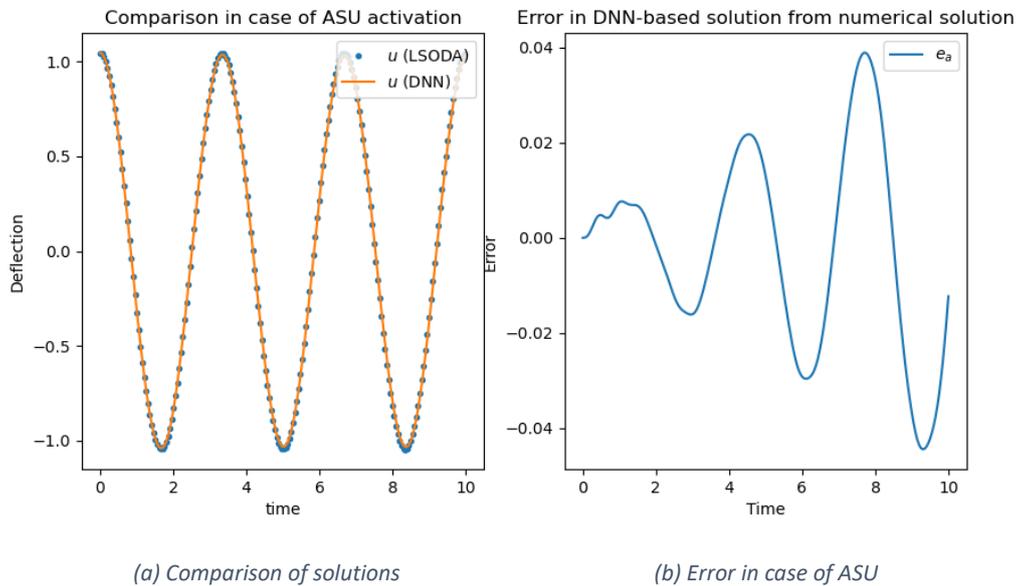

(a) Comparison of solutions

(b) Error in case of ASU

**Fig. 15** *Comparison of solutions from both computational techniques and the difference (error) of DNN-based solution from Numerical approximation*

Therefore, to study the dynamics of nonlinear periodic systems using DNNs, the choice of oscillatory activation functions is more radical and effective with faster training cum less computational cost. One can have a glance on Table 2 to get an idea about the performance and effectivity of different activation functions being discussed.

**Table 2** Summary of training DNNs to deal the modeled MEMS with different activation functions

| Activation function | Training epochs | Training time |
|:---:|:---:|:---:|
| Tanh | 35,000 | 18 minutes 33 seconds |
| Mish | 20,000 | 17 minutes 50 seconds |
| Sine | 6,000 | 3 minutes 16 seconds |
| GCU | 4,000 | 2 minutes 53 seconds |
| ASU | 2,000 | 1 minute 14 seconds |

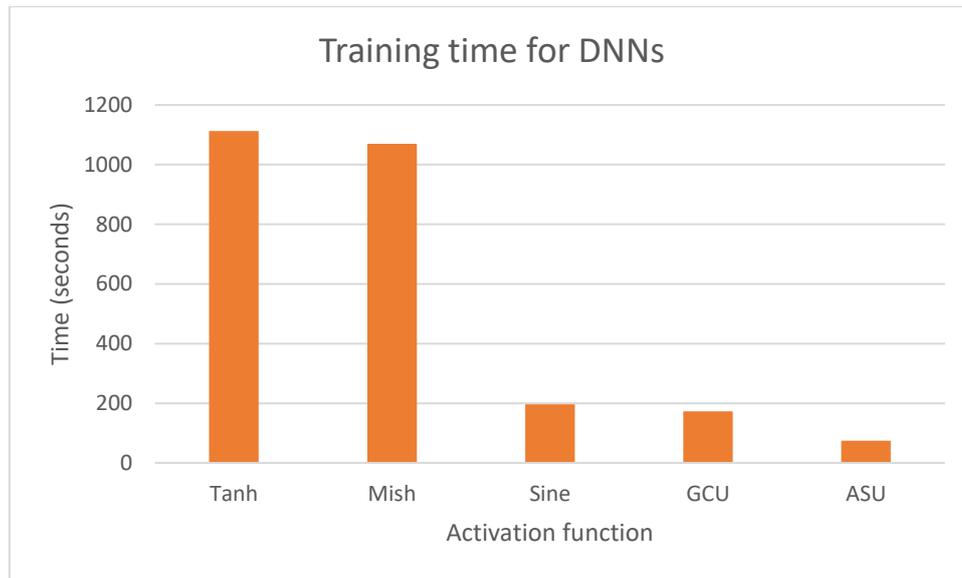

*Fig. 16    Time taken by each DNN architecture with different activation functions to simulate the solution of MEMS undergoing electric actuations*

Fig. 16 endorses the use of oscillatory activation functions for the cases of handling nonlinear periodic behaviors of systems. It also confirms the computational feasibility and effectiveness of proposed ASU to deal the complex, periodic and nonlinear dynamical systems.

**6 Conclusion**

This work provides a comprehensive analysis on the behaviors of different oscillatory and non-oscillatory activation functions regarding highly nonlinear and periodic systems. Experimental results clearly approved the computational feasibility, effectivity and robustness of oscillatory activation functions for handling pure periodic behaviors. The performance of integrated GCU with DNNs for MEMS is better then Sine, Tanh and Mish. The proposed ASU outperformed GCU in predicting the nonlinear periodic behaviors of electrically actuated microbeam under consideration. As a future work, the role of proposed activation function can also be examined on different tasks with other deep network structures, loss functions and optimization techniques.